\documentclass{article}

\usepackage{arxiv}

\usepackage[utf8]{inputenc} % allow utf-8 input
\usepackage[T1]{fontenc}    % use 8-bit T1 fonts
\usepackage{hyperref}       % hyperlinks
\usepackage{url}            % simple URL typesetting
\usepackage{booktabs}       % professional-quality tables
\usepackage{amsfonts}       % blackboard math symbols
\usepackage{nicefrac}       % compact symbols for 1/2, etc.
\usepackage{microtype}      % microtypography
\usepackage{graphicx}
\usepackage[numbers]{natbib}
\usepackage{doi}

\usepackage{algorithm}
\usepackage[noend]{algpseudocode}
\floatname{algorithm}{Algorithm}
\usepackage{amsmath,amssymb,amsfonts}%
\usepackage{amsthm}%
\usepackage{mathrsfs}%
\usepackage{tabularx}%
\usepackage{array}
\usepackage{booktabs}
\usepackage{tabularx}
\usepackage{array}
\usepackage{threeparttable}

\usepackage[utf8]{inputenc}  
\usepackage{listings}        
\usepackage{xcolor}          
\title{Robot guide with multi-agent control and automatic scenario generation with LLM}

\author{ \href{https://orcid.org/0009-0005-8580-2567}{\includegraphics[scale=0.06]{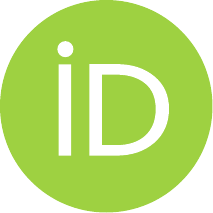}\hspace{1mm}Elizaveta D.~Moskovskaya} \\
	National Research Center\\ 
    "Kurchatov Institute"\\ 
    Moscow, Russia\\
	\texttt{moscovskayaliza@gmail.com} \\
	\And
	\href{https://orcid.org/0000-0002-6546-8697}{\includegraphics[scale=0.06]{orcid.pdf}\hspace{1mm}Anton D.~Moscowsky} \\
	National Research Center\\ 
    "Kurchatov Institute"\\ 
    Moscow, Russia\\
	\texttt{moscowskyad@gmail.com} \\
}

\renewcommand{\shorttitle}{Robot guide with multi-agent control and automatic scenario generation with LLM}

\hypersetup{
    pdftitle={Robot guide with multi-agent control and automatic scenario generation with LLM},
    pdfsubject={cs.RO, cs.AI, cs.HCI},
    pdfauthor={Elizaveta D.~Moskovskaya, Anton D.~Moscowsky},
    pdfkeywords={Multi-agent control system, LLMs, Robot tour guide, Scenarios},
}

\begin{document}
\maketitle

\begin{abstract}
	The article describes the development of a hybrid social robot control architecture to overcome the limitations of traditional approaches, where behavior scripts manually synchronize the robot's actions and text, and existing methods focus primarily on short dialogue responses. The architecture of the proposed system combines a multi-agent resource management system with automatic generation of behavior scenarios based on large language models. This system automates the preparation of text and commands for the robot's non-verbal behavior for extended narratives and resolves resource conflicts between multiple execution mechanisms. The system was tested on the MENTOR-1 tour guide robot, for which it successfully generated scenarios automatically and demonstrated more natural and rich behavior compared to existing approaches.The proposed approach provides full automation of both scenario preparation and execution through efficient resource management, enhancing the quality of social robot interaction in long-term storytelling tasks.
\end{abstract}

% keywords can be removed
\keywords{Social robot \and  Multi-agent control system \and LLM \and Robot tour guide \and Scenarios}

\section{Introduction}
Anthropomorphic social robots are now widely applied in education \cite{Belpaeme2018}, healthcare \cite{Silvera-Tawil2024} and the service industry \cite{Gonzalez-Aguirre2021} (e.g., promotion and guided tours). During operation, such robots interact with humans via speech. But a critical element of effective human-machine interaction is the use of nonverbal communication, including gestures and eye contact. A body of experimental research demonstrates that robots employing nonverbal communication are evaluated more positively by humans. For example, pointing gestures influence human attention focus  \cite{Zinina2020}, while eye contact positively affects human subjective evaluations of the robot \cite{Zinina2024, Abdollahi2022}. Studies on human behavior \cite{Rodero2022} also highlight the impact of nonverbal gestures and their frequency on human-human interaction. Importantly, the mentioned nonverbal communication modalities must be synchronized with the robot's speech. This synchronization can be achieved through pre-scripted scenarios created manually. However, the need to enhance the autonomy of anthropomorphic robots - including the ability to communicate in natural language on arbitrary topics - motivates the development of systems that automatically accompany arbitrary spoken text with a set of actions. Large language models (LLMs), which enable robots to interact with humans in natural language, can also be adapted to generate accompanying actions processed by the control system. 

This work proposes a system that, using a tour-guide robot as an example, generates a narrative about an exhibit based on its extended description and accompanies it with a series of nonverbal actions, including gestures and limb animations, as well as facial expressions and eye contact. Narrative generation and action integration are performed in two stages using a base (non-fine-tuned for the specific task) LLM. The scenario obtained from the LLM is processed by a developed multi-agent control system, which resolves potential conflicts in robot resource allocation and temporal overlaps of individual actions, while also addressing transitions back to baseline behavior. As the terms "action", "behavior", "scenario", etc. lack universally accepted definitions, the authors specify their meanings in the context of this work (Table~\ref{tab:definitions}).

\begin{table}[htbp!]
\centering
\caption{Author definitions of some terms used in work}
\label{tab:definitions}
\begin{tabularx}{\linewidth}{>{\raggedright}p{2.25cm} >{\raggedright\arraybackslash}X >{\raggedright\arraybackslash}X >{\raggedright\arraybackslash}X >{\raggedright\arraybackslash}X}
    \toprule
    \textbf{Term} & \textbf{Definition} & \textbf{Example} \\
    \midrule
    Action & Elementary unit of robot interaction with the environment & Gestures and limb animations, speech synthesis (narratives), demonstration of emotions via facial expressions, etc. \\
    \addlinespace
    Behavior & Sequence of actions to achieve a specific goal taking sensors data in account & Establishing and maintaining eye contact, moving to coordinates while avoiding collisions \\
    \addlinespace
    Scenario/Script & Sequence of events, conditions, and goals within which the robot performs one or more actions and behaviors & Intro and outro speeches, exhibit presentment \\
    \addlinespace
    Gesture & Action resulting in the robot changing the position of joints in its limbs (either one or several simultaneously)	& Pointing to oneself, frightened pose (holding his head with his hands), speaker's posture (one hand on the "heart", the other extended towards the audience) etc. \\
    \addlinespace
    Animation & Action comprising a fixed sequence of gestures & Greeting hand wave, head shaking, nodding etc. \\
    \addlinespace
    Eye contact	& Behavior that positions the robot's head and pupils such that they "look" at the recognized eyes of the user &	Contact with the tour audience \\
    Pointing gesture & Behavior directing one or both of the robot's arms, as well as the head and pupils, toward a point in space; also performs a full body turn if required	& Pointing to an object \\    
    \addlinespace
    Facial expression & Action setting the movable elements of the face (such as brows and pupils) in a specific configuration &	Joy, sadness, neutral \\
    %\addlinespace
    %(Action) tag & An action command for a robot (in the scenario), in a special format, containing the type of action to be performed and its parameters & Tag to set the facial expression to a sad emotion: \textless facial:sad \textgreater \\
    \bottomrule
\end{tabularx}
\end{table}

\section{Review}
Accompanying the robot's speech with a set of nonverbal actions can be specified via a rigid script indicating which action to perform at each moment of the narrative. Researchers often develop custom languages for this purpose, as in \cite{AlvaradoVasquez2020}. Since actions typically have some duration, specialized graphical user interfaces may be used for visual clarity in the generation of such scripts \cite{Bodunkov2020}. Despite the possibility of iterative debugging, these scenarios are limited to narrow domains and are unsuitable for robots engaged in dialogue on arbitrary topics.

Several researchers propose generating the robot's pose and gestures using a fixed set from specially trained neural network architectures. For example, \cite{Yoon2019, Yoon2020Speech} and \cite{Li2024} employ recurrent neural networks to generate direct control of robot arm joints without semantic labels for individual actions. These architectures are trained on video datasets of speaking and gesticulating humans, with performance heavily dependent on whether public speakers are used, as in \cite{Yoon2019}, or everyday human interactions, as in \cite{Li2024}. The work \cite{Teshima2022} also addresses speech-based gesture generation, but categorizes gestures into three classes: beat gestures, imagistic gestures, and no gesture. In \cite{Tuyen2023}, a similar paradigm is followed, which generates the full pose of the robot while considering the pose and speech of the human interlocutor. Likewise, \cite{VallsMascaro2025} generates robot poses using a transformer model based on the interacting human's pose and a textual description of the desired interaction. Although these works enable coherent actions, they are highly dependent on training data and lack full spatial grounding required for actions like pointing at objects or eye contact.

To link reproducible text uttered by the robot with actions, it is preferable to use a library of such actions selected by semantic relevance. Behavior trees (BTs) \cite{Iovino2022} enable the integration of various actions in robot control. Originally created as a developer tool, BTs have also been constructed using LLMs' programming capabilities. The papers \cite{Lykov2024} and \cite{Izzo2024} describe fine-tuning LLMs to build BT from textual robot instructions. However, these works do not focus on social robots, which have unique requirements for text-action synchronization, and rely on predefined node libraries describing the specific robot's actions-incorporating arbitrary text segments requires separate development. 

Several robotics works address planning actions via LLMs from textual instructions. Approaches like SayCan \cite{Ahn2022}, PaLM-E \cite{Driess2023}, Code as Policies \cite{Liang2023} and STRL \cite{Mironov2024} form action chains from available primitives leading to a final goal. These consider household mobile-manipulator robots, whereas accompanying a social robot's narrative with nonverbal actions lacks an explicit goal for sequencing actions, as nonverbal actions merely enhance human-robot interaction effects without being strictly necessary.

Another LLM-based approach selects the most suitable action from an available set to match the uttered text. \cite{Wang2024} addresses emotional prosody in spoken text, but not in nonverbal actions. \cite{Lee2023} selects an action from a list based on both the robot's spoken phrase and the user's query in dialogue, supporting hand gestures, facial expressions, and special actions such as the establishment of eye contact. However, the system processes isolated 1-2 sentence phrases, selecting only one action and precluding combinations like gesticulating during eye contact. \cite{Kim2024} tackles a similar task for a virtual agent rather than a physical robot: an action is selected from a list that best matches the agent's phrase and its emotional tone, with a separate module generating control signals synchronized in rhythm with the audio. Yet, only one action per phrase is supported, and while eye contact with the user is mentioned, details on interactions with head-involving animations (potentially causing conflicts) are absent. \cite{Galatolo2025} proposes a similar approach, delegating phrase segmentation and action selection from a list to a language model-using a small model deployable locally on the robot's computer. Experiments show multiple non-conflicting actions selectable per phrase (e.g., eye and eyebrow movement), but action conflict resolution is not addressed. \cite{Kang2024} proposes a system in which LLM outputs an unbounded list of actions in JSON format. The humanoid robot Nadine supports extensive capabilities, including gestures, animations, emotions, and gaze control (including eye contact). The system ensures smooth transitions between sequentially selected emotions, including resets to default states, via a dedicated animation engine that blends segments for smoothness. However, the engine's operating principle and its ability to resolve temporal overlaps or resource conflicts remain unspecified.

Table~\ref{tab:review} presents the mentioned works that focus on the automatic accompaniment of robot speech with nonverbal actions. However, the reviewed studies do not emphasize the problem of resolving conflicts between different actions when they involve shared resources (for example, both actions engage the robot's head). The objective of this work is to develop a system capable of automatically resolving these conflicts, which will also enable more saturated nonverbal accompaniment of the robot's narrative with actions-positively impacting user impressions, as supported by research in human-machine interaction \cite{Zinina2020, Zinina2024, Abdollahi2022} and psychology \cite{Rodero2022}.

\begin{table}[htbp!]
\centering
\caption{Works on Using Language Models for Accompanying Speech with Nonverbal Communication Means}
\label{tab:review}
\begin{tabularx}{\linewidth}{>{\hsize=0.7\hsize\raggedright\arraybackslash}X 
                            >{\hsize=1.9\hsize\raggedright\arraybackslash}X 
                            >{\hsize=0.7\hsize\raggedright\arraybackslash}X 
                            >{\hsize=0.7\hsize\raggedright\arraybackslash}X}
    \toprule
    \textbf{Article} & \textbf{Actions and behaviors} & \textbf{Parallel execution} & \textbf{Conflict resolution} \\
    \midrule
    \cite{Wang2024} & Facial expressions & \texttimes{} & \texttimes{} \\    
    \addlinespace
    \cite{Lee2023} & Facial expressions, gestures and animations, eye contact & \texttimes{} & \texttimes{} \\ 
    \addlinespace
    \cite{Kim2024} & Gestures and animations, eye contact &  \texttimes{} & \texttimes{} \\ 
    \addlinespace
    \cite{Galatolo2025} & Facial expressions, gestures and animations &  \checkmark & \texttimes{} \\ 
    \addlinespace
    \cite{Kang2024} & Facial expressions, gestures and animations, gaze control &  \checkmark & ? \\ 
    \addlinespace
    This article approach & Facial expressions, gestures and animations, gaze control, body rotation, interaction with objects & \checkmark & \checkmark \\
    \bottomrule
\end{tabularx}
\end{table}

\section{The MENTOR-1 approach}\label{secmentor}
\begin{figure}[htbp!]
    \centering
    \includegraphics[width=\linewidth]{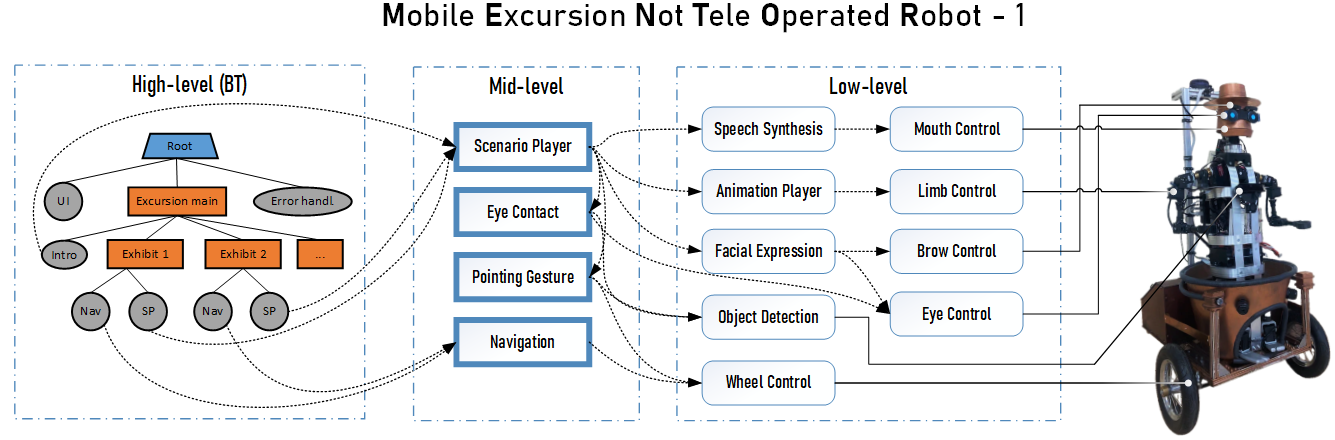}
    \caption{MENTOR-1 simplified architecture}
    \label{fig:robot}
\end{figure}

\noindent To test means of automatic accompaniment of a robot's narrative with nonverbal actions, the MENTOR-1 robot (Fig. \ref{fig:robot}) was employed, developed at the NRC "Kurchatov Institute" for conducting tours of the robotics laboratory \cite{Malyshev2023}. The robot features an anthropomorphic design, with movable limbs for nonverbal gestures and a head capable of displaying various facial expressions through movable eyebrows and display eyes. The robot's control system has a classical multi-level structure within the ROS framework \cite{Quigley2009}. At the low-level, individual nodes control robot elements: motors, limbs, face, speech, sensor data collection, processing, etc. The mid-level implements individual behaviors, including navigation to a target, establishing eye contact by detecting human faces, performing pointing gestures, etc. Additionally, at the middle level resides a scenario player module, with scenarios defined in a special format as robot's speech text with embedded tags that trigger action execution:

\begin{verbatim}
Some text pronounced by robot <action_type:parameter_1; … > ...
\end{verbatim}

These scenarios describe the robot's behavior near exhibits or during greetings and farewells, i.e., specific behaviors in a specific location. Scenarios primarily involve textual communication, gestures, animations, and facial expressions, including establishing eye contact. At the high-level, there is a controlling Behavior Tree that forms a specific tour, including introductory phrases, movement between exhibits, and launching scenarios that present them, as well as human control and recovery mechanisms in case of errors.
The specified scenario format is convenient because it intuitively allows one to synchronize the start of action execution with the spoken text by adjusting the tag position in the text, and it is sufficiently simple for manual editing even by individuals without programming skills. However, the inconvenience of these scenarios lies in the difficulty of assessing the completion of the action, leading to conflict situations where simultaneous requests for actions that utilize shared resources occur. There are some examples of such conflicts, that was faced on MENTOR-1:
\begin{itemize}
    \item Different gestures/animations involves same arm.
    \item Some animation used head, that already involved in eye-contact.
    \item Robot performs pointing gesture during other animation.
\end{itemize}
When, in case of conflict, priority was given to the newly arrived action, this caused issues with movement smoothness; if the older action had priority, it impoverished the robot's performance, making it stationary too long. Additionally, forming a sequence of gestures or facial expressions required configuring the robot's return to a default pose/expression. If the scenario composer forgot to "reset" such state to default, the robot retained an inappropriate pose/expression until the next action.

These problems were partially addressed through manual tuning with result evaluation via simulation or on the robot, which required several iterations. The other problem within the task of a tour-guide robot is the complete repeatability of tours. All the described problems can be solved by creating a tool that automatically generates unique and personalized narratives about exhibits and automatically accompanies them with actions and behaviors from the robot's available library, while resolving conflict issues and returning to default behavior.

\section{Suggested Solution}

The following requirements were set before the developed solution:

\begin{enumerate}{}{}
\item{Automatic generation of narratives about exhibits, taking into account various settings, including duration, target audience, and style.}

\item{Automatic formation of a scenario based on the narrative obtained by adding actions and behaviors from an extensive library.}

\item{Execution of the obtained scenarios, resolution of conflict issues, and return to default behavior. Return to default behavior is implied both during the scenario (during periods of inactivity) and upon completion of the scenario (pose for safe movement between exhibits).}
\end{enumerate}

The proposed solution replaces the scenario player block in the original architecture (Fig. \ref{fig:robot}). At the same time, it was decided to retain the previous scenario format in the form of spoken text with embedded special action tags, which allows the use of an LLM for working with it without additional modifications.

So, the solution consists of two parts: \textbf{a scenario generation module} and \textbf{a multi-agent control system}.

The scenario generation process consists of two stages; the first stage involves compressing and stylizing an extended exhibit description into a coherent narrative adapted to the robot's storytelling style and the type of audience. The second stage involves placing action tags within the narrative according to the context.

\begin{figure*}[htbp!]
    \centering
    \includegraphics[width=\linewidth]{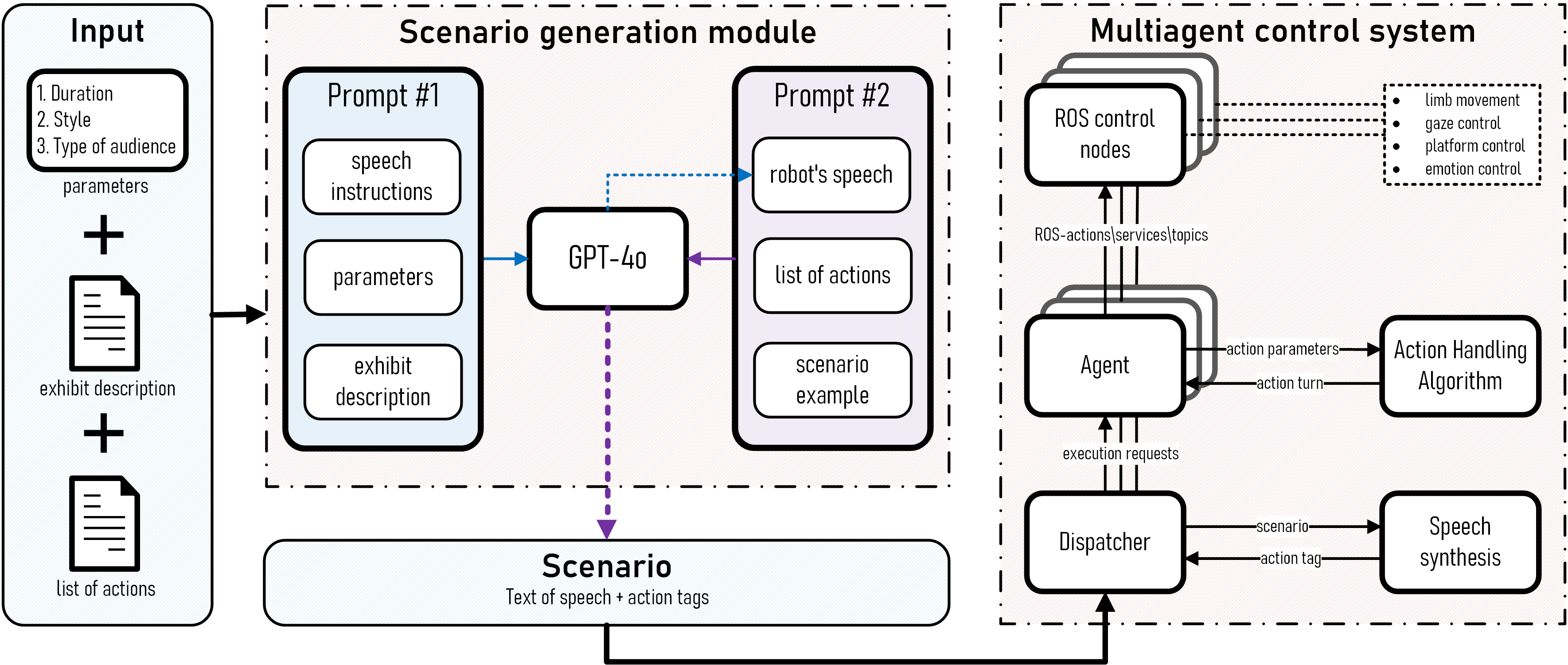}
    \caption{The process of creating and implementing an excursion scenario.}
    \label{fig:fig1}
\end{figure*}

When generating narratives using LLM, in addition to mentioned synchronization, conflict and reset-to-default issues, hallucinations may occur\cite{Huang2025}, which require the system to self-repair after incorrect commands/parameters. The solution to all four of these problems lies at the control system level. The control system is based on a multi-agent approach. This approach has proven its effectiveness in the context of managing both multiple robots \cite{Chen2021} and one \cite{Amaike2024, Costantini2019}. In this work, multi-agent systems are also considered within a single physical robot, where each agent responsible for a specific robot part has a default behavior/action to which it strives to return after completing its task action. Some agents also have a list of background actions performed during extended periods of inactivity. Agents are organized in a hierarchical structure, where each action is assigned a priority level. The interaction between agents is regulated using the priority of the action and the position of the agent in the overall hierarchy. Based on these two characteristics, the system determines whether to execute, cancel, postpone, delegate, or modify an action. This conflict resolution mechanism (for situations where multiple actions require the same resources) ensures flexible action execution.
Agents receive commands to perform actions via a dispatcher, which processes scenario instructions and sends execution requests to the specific agent. 
The complete pipeline of the proposed solution is shown in Fig.~\ref{fig:fig1}.

\section{Scenario generation module}\label{secgen}
\noindent The scenario generation module is implemented as a ROS node. It is designed for the automatic creation of behavior scenarios to narrate each exhibit in a tour based on its extended description. Generation proceeds in two steps:

\begin{enumerate}{}{}
\item{A stylized narrative is created from the perspective of the robot guide.}
\item{Nonverbal action tags, such as gestures and emotions, are inserted into the resulting narrative.}
\end{enumerate}

The robot saves and uses the final scenario to carry out the tour. In the absence of a connection, the robot will play one of the previously generated scenarios that is closest in parameters. In addition, the robot always has access to a basic scenario created manually.

OpenAI GPT models \cite{Islam2024} are used for generation, interacting via requests that contain instructions for creating natural language scenarios and the description of the exhibit. The request specifies generation parameters: narrative length, tour style (formal, humorous), and audience type (schoolchildren, adults without technical background or specialists). The instructions include information about the location, the narrator, the generation parameters, the presentation rules, and the interaction with the audience.

The action tag placement process enriches the scenario with nonverbal signals: pointing on exhibits, facial expressions, gestures, and animation playback. The instructions provide the models with a list of available actions, rules for their use, and contextual meaning. Usage rules include the format for forming the tag, as well as recommendations for placement. Instantaneous actions are triggered immediately before the corresponding text, while longer actions are triggered in advance, which affects their placement.

Thus, the proposed method ensures flexible scenario generation, automating the tour creation process and minimizing the need for specialist involvement. However, it should be understood that such generation inevitably leads to situations involving action conflicts and incorrect commands. These issues are resolved at the control system level.

\section{Control system}\label{seccon}
The multi-agent system is based on the principle of modularity, which makes it extensible and capable of solving motion coordination tasks. The system architecture includes two main components: the dispatcher and the agent forest.

Within this architecture, an \textbf{agent} is the fundamental autonomous unit of control and execution. Each agent:

\begin{itemize}
    \item \textbf{Represents} a specific functional or physical component of the robot (e.g., a limb, head, facial expression subsystem, or a coordinator of a behavior like establishment of eye contact).

    \item \textbf{Is Responsible} for managing a defined set of actions (e.g., moving joints, setting facial expression, tracking audience faces) and maintaining its internal state.

    \item \textbf{Interacts} hierarchically. Agents are organized in trees where a parent agent can delegate tasks to its children and manage conflicts based on priority and hierarchical position. Agents within the same tree coordinate resource access and action execution, when agents in separate trees (e.g. responsible for arms and brows) operate independently.

    \item \textbf{Operates Autonomously}. Agents encapsulate their behavior logic, handle action execution, conflict resolution, automatic return to default behavior during inactivity, and execute background actions.
\end{itemize}

The \textbf{dispatcher} is an ROS node responsible for initializing agents, defining their hierarchy, and transmitting commands to agents of the scenario. The dispatcher does not control agents directly; it only forwards execution requests and handles cancellations and the global state of agents, enabling coordination of the entire system.

An abstract agent class was created as the foundation for all specialized agents. This class implements action management mechanisms, a conflict resolution algorithm, automatic return to default behavior, and the execution of background actions during inactivity. 
For agents with predefined default behavior, a timer is used to track the time since the last successful operation. When the threshold is exceeded, a default behavior is triggered with a minimum priority.
The logic is similar to background actions, if the agent remains in the default behavior for more than a certain amount of time (and has a predefined list of background actions), a random action is performed from the list, with a lower priority than any other action except the default behavior.
An important feature is the hierarchical organization of agents, where each agent can have one parent and multiple children, forming a tree structure that resolves resource access conflicts based on priorities and hierarchical position. Some agents could be fully independent of each other, like agents controlling the eyebrows and arms, so they are placed in separate trees. This organization ensures that higher-priority actions or those belonging to agents higher in the hierarchy can cancel or block lower-priority actions.

\begin{figure}[htbp!]
    \centering
    \includegraphics[width=\linewidth]{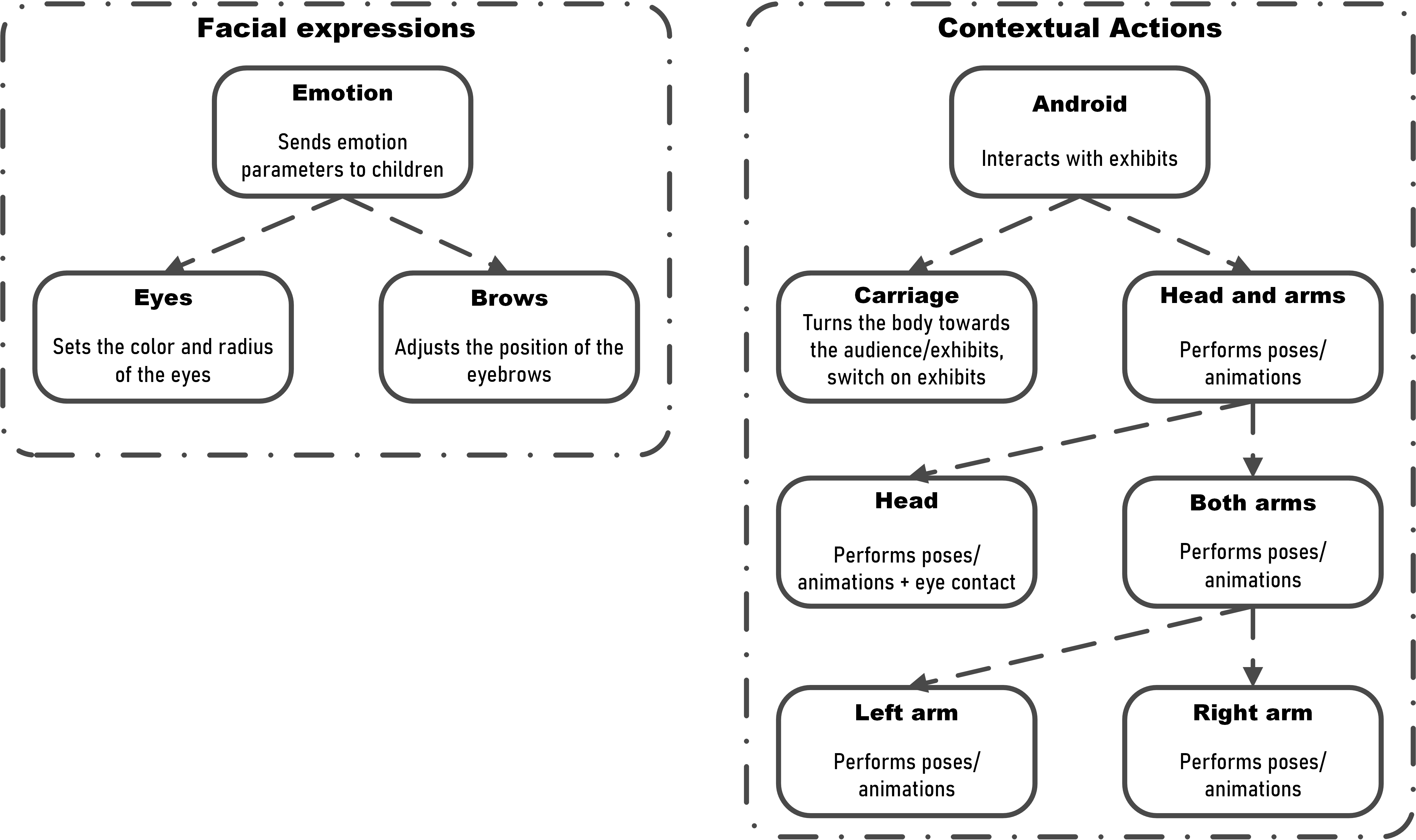}
    \caption{The hierarchy forest of agents used to control the MENTOR-1 robot}
    \label{fig:fig2}
\end{figure}

For the robot guide, a hierarchy of agents was created. It includes several limb agents as well as agents for the head, emotions, torso, and a top-level agent for complex actions. The complete hierarchy as a forest is shown in Fig.~\ref{fig:fig2} and the list of agents with the list of actions available to them in Table ~\ref{tab:agents_actions}.

\begin{table}[htbp!]
\centering
\begin{threeparttable}
\caption{Summary of agents, their available actions by command, and default behavior}
\label{tab:agents_actions}
\begin{tabularx}{\linewidth}{>{\raggedright}p{2.5cm} >{\raggedright\arraybackslash}X >{\raggedright\arraybackslash}X}
    \toprule
    \textbf{Agents} & \textbf{Available actions by command} & \textbf{Default behavior} \\
    \midrule
    Right arm, Left arm, Both arms, Head and arms & Set joint positions; perform animation & Set limbs to neutral pose \\
    \addlinespace
    Head & Set joint positions; perform animation; establish gaze contact with a 3D point & Recognizing faces and making eye contact with a random listener \\
    \addlinespace
    Carriage & Rotate torso towards a 3D point & Rotate torso towards the center of the audience group \\
    \addlinespace
    Android & Interact with exhibit (rotate, look at and point on it) & None\tnote{*} \\
    \addlinespace
    Emotion & Set emotional expression & Set neutral facial expression \\
    \addlinespace
    Brows & Set eyebrows position & None\tnote{*} \\
    \addlinespace
    Eyes & Set emotional tint of eyes & None\tnote{*} \\
    \bottomrule
\end{tabularx}
\begin{tablenotes}
\small
\item[*] (it has no personal default behavior, but it can be used as part of an ancestor or a descendant behavior).
\end{tablenotes}
\end{threeparttable}
\end{table}

The limb agents manage poses and animations of various joints, such as the left and right arms and their combinations, enabling synchronization of complex movements. The Head agent extends limb functionality by adding control of gaze and face tracking of listeners, implemented through sequential smooth eye movements and head turns. The Emotion agent is responsible for the emotional non-verbal coloring of speech, synchronizing eyebrow position, and eye color. The Carriage agent manages spatial orientation of the robot, ensuring a turning toward the audience or exhibit. The top-level Android agent handles interaction with the exhibit by coordinating the actions of other agents, including turning to the exhibit, pointing at objects (with appropriate arm selection) and establishing eye contact.

To enhance the smoothness and naturalness of the robot’s movements, several conflict resolution policies for overlapping tags were implemented. In the context of speech-synchronized actions, resuming an interrupted gesture after executing a higher‑priority one would often result in unnatural timing, as the original action is no longer relevant to the ongoing narration. Therefore, the system cancels the current action when a new one with higher priority arrives, ensuring that the robot’s nonverbal behavior always aligns with the immediate content of the speech. The complete logic is shown in Algorithm~\ref{alg:agent_handling}.

\begin{algorithm}
\caption{Action Handling Algorithm}
\label{alg:agent_handling}
\begin{algorithmic}[1]
\Statex \textbf{Input:} 
\Statex \quad New action $a$ with parameters
\Statex \quad Optional override priority $P_{\text{override}}$
\Statex \quad Flag $\text{from\_parent}$: indicates if action is delegated from direct ancestor
\Statex \quad Agent: 
\Statex \quad \quad State: $\text{active}$ or $\text{stopped}$
\Statex \quad \quad Current action: $a_c$ (if any) with priority $P_c$
\Statex \quad \quad Parent: link to parent or None
\Statex
\Statex \textbf{Notations:}
\Statex \quad $P_{a} \gets \begin{cases} 
                          P_{\text{override}} & \text{if provided} \\
                          \text{$a$.base\_priority} & \text{otherwise}
                        \end{cases}$
\Statex \quad $P_c \gets \text{current action priority}$
\Statex \quad $P_{\text{anc}} \gets \text{max priority in ancestor hierarchy}$
\Statex \quad $P_{\text{des}} \gets \text{max priority in descendant hierarchy}$
\Statex \quad $\text{cancel}(X) \equiv \text{cancels the execution of } X$

\If{agent is $\text{stopped}$}
    \State \Return $\text{ignore}$
\EndIf

\If{parent exists \textbf{and not} $\text{from\_parent}$}
    \If{$P_{a} > P_{\text{anc}}$ \textbf{and} $P_{\text{anc}} \neq 0$}
        \State $\text{cancel}(\text{ancestors' actions})$
    \Else
        \State \Return $\text{ignore}$
    \EndIf
\EndIf

\If{$P_c < P_{a}$}
    \If{$P_{\text{des}} < P_{a}$}
        \State $\text{cancel}(\text{descendants' actions})$
    \EndIf
    
    \State $\text{cancel}(\text{agent's current action})$
    \State Set agents' current action to $a$ and $P_c \gets P_{a}$
     \State $\text{execute}(a)$
\Else
    \State \Return $\text{ignore}$
\EndIf
\end{algorithmic}
\end{algorithm}

For all agents that implement long-running actions, an adapted version of the main conflict resolution algorithm is used, which delays the start of a new action if the current action is more than 75\% complete, and otherwise implements the basic logic. The top-level Android agent complements the basic conflict resolution algorithm by assigning mirror actions to perform pointing gestures with a limb that is occupied by a lower-priority action.

The system implements an automatic return to default behavior. Background actions are also performed during extended inactivity, maintaining action richness during periods without explicit commands. The proposed control system, combined with the scenario generation system, allows the automatic creation of unique and parameterized tour scenarios for the robot, completely eliminating the need for manual tuning.

The implementation is available on GitHub \footnote{Repository: \url{https://github.com/moskovskayaliza2002/multiagent_system}.}.

\section{Experiments}
The purpose of the conducted experiments is to determine the effectiveness of the developed multi-agent approach in comparison with existing approaches which generate nonverbal behavior for social robots focused on short dialogue utterances. Such scenarios require a limited number of actions and are typically confined to one or two non-conflicting behaviors \cite{Galatolo2025}, \cite{Lee2023}. However, in tasks that involve extended narration, the robot must simultaneously manage multiple resources: facial expressions, gesticulation, and torso movements. This requires the implementation of action prioritization and conflict resolution mechanisms, which are absent in current systems.

\subsection{Baseline system}

A baseline system was implemented, adopting the approach from \cite{Galatolo2025}, \cite{Lee2023}. By using LLM it assigned exactly one emotion tag and one other action tag per sentence from the robot's overall action list. The baseline system's action list encompassed all actions from the multi-agent system, plus default behaviors in the multi-agent system (turning toward listeners, establishing eye contact, adopting a neutral pose). The annotation prompt was structured similarly: it used the same descriptions of contextual actions-including new actions, examples, and rules; but adapted to the baseline system's annotation format:

\begin{enumerate}{}{}
\item{Exactly one emotion tag per sentence,}
\item{Exactly one other action tag per sentence,}
\item{After some actions marked as prominent, prefer an action returning to neutral behavior unless the next sentence also contains a prominent action.}
\end{enumerate}

\begin{figure*}[t!]
    \centering
    \includegraphics[width=\linewidth]{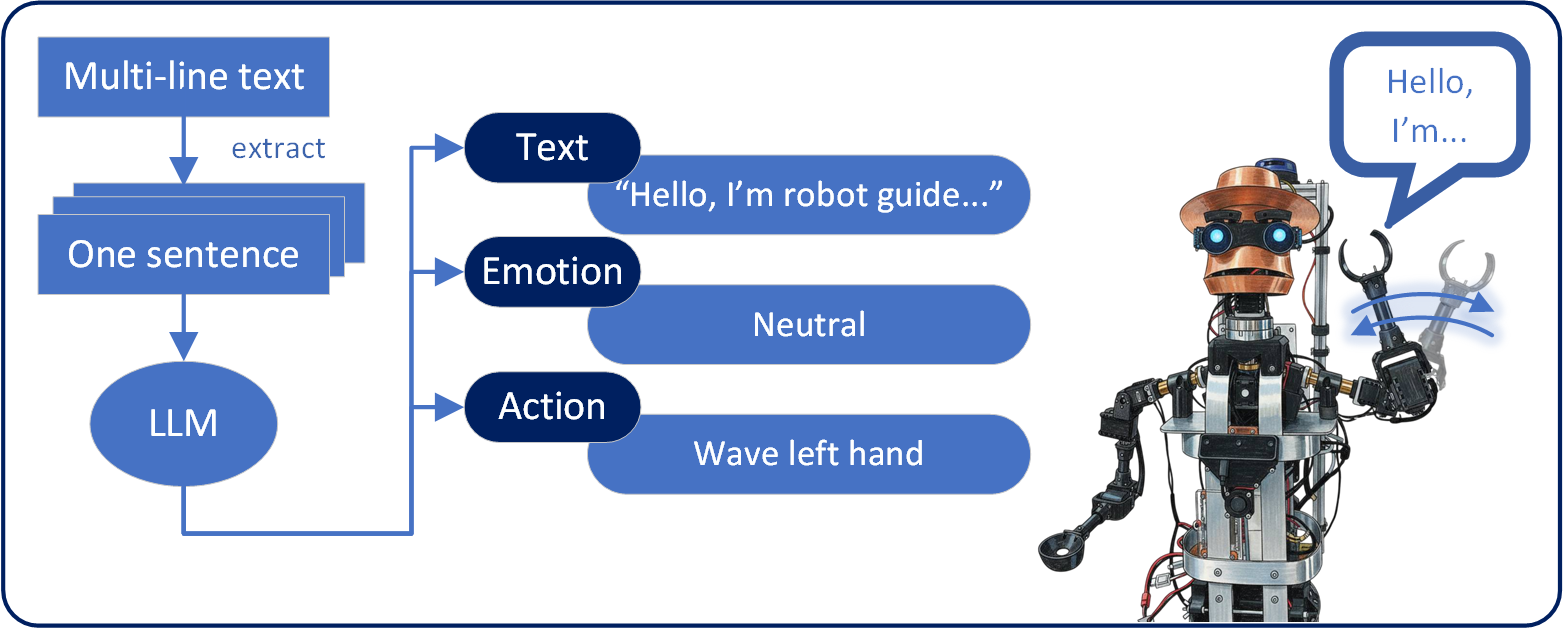}
    \caption{Scheme of the implemented baseline system's operation principle.}
    \label{fig:fig3}
\end{figure*}

Some action/behavior execution may be longer than speech of its sentence, so the next conflict resolution mechanism was added. A new action interrupts all ongoing actions that utilize at least one resource required by the new action. The general principle of operation of the baseline system is shown in Fig. ~\ref{fig:fig3}.

\subsection{Setup and metrics}

The experiment was conducted using the MENTOR-1 (section \ref{secmentor}) tour guide robot, where the block scenario player at Fig. \ref{fig:robot} was changed in the baseline and multi-agent systems. 14 scenarios that describe 14 exhibits were identical for both systems, and the same version of LLM (GPT-4o) was used to generate supporting actions. The descriptions of available actions were also identical. 
The multi-agent system executed the scenarios following all aforementioned rules (see sections \ref{secgen} and \ref{seccon}).

During each scenario execution, the following metrics were computed:
\begin{enumerate}{}{}
\item{\textbf{The number of interrupted actions} is a classical metric for action planning systems, reflecting the capacity for resource coordination.}
\item{\textbf{The total number of executed actions} reflects nonverbal behavior density, studies indicate that a higher frequency of gestures enhances listener attention and engagement.}
\item{\textbf{The proportion of non-neutral body actions and emotions} reflects the expressiveness of nonverbal signals.}
\item{\textbf{Idle time of the limb}; effective interaction requires various nonverbal signals, thus necessitating activity distribution across robot parts.}
\item{\textbf{Silent animation}; captures nonverbal activity during speech pauses.}
\end{enumerate}

\subsection{Results and discussion}
Table \ref{tab:results} shows the average metrics values for 13 scenarios for both systems. The relative change is calculated as:
\[ \frac{new - old}{old} \times 100\%, \]
\noindent Where \textit{new} stands for multi-agent system values, \textit{old} stands for baseline system values.

\begin{table}[htbp]
\centering
\caption{Average metrics value for each system}
\label{tab:results}
\begin{tabularx}{\linewidth}{>{\hsize=1.9\hsize\raggedright\arraybackslash}X 
                            >{\hsize=0.7\hsize\raggedright\arraybackslash}X 
                            >{\hsize=0.7\hsize\raggedright\arraybackslash}X 
                            >{\hsize=0.7\hsize\raggedright\arraybackslash}X}
    \toprule
    \textbf{Metric, unit of measurement} & \textbf{Baseline system} & \textbf{Multi-agent system} & \textbf{Relative change, \%} \\
    \midrule
    Scenario duration, s & 65.43 & 69.44 & +6.1 \\
    \addlinespace
    Interrupted actions, pcs, \textdownarrow & 0.92 & 0.23 & –75.0\\
    \addlinespace
    Total actions, pcs, \textuparrow & 16.77 & 25.31 & +50.9\\
    \addlinespace
    Body (non-default) actions, pcs, \textuparrow & 5.38 & 6.77 & +25.7 \\
    \addlinespace
    Body (default) actions, pcs, \textuparrow & 2.69 & 5.54 & +105.7 \\
    \addlinespace
    Emotion (non-default) actions, pcs, \textuparrow & 2.15 & 3.08 & +42.9 \\
    \addlinespace
    Emotion (default) actions, pcs, \textuparrow & 6.54 & 10.00 & +52.9 \\
    \addlinespace
    Right-hand idle time, \%, \textdownarrow & 88.85 & 73.65 & -17.1 \\
    \addlinespace
    Left-hand idle time, \%, \textdownarrow & 88.33 & 74.25 & -15.9 \\
    \addlinespace
    Head idle time, \%, \textdownarrow & 85.08 & 83.61 & -1.7 \\
    \addlinespace
    Carriage idle time, \%, \textdownarrow & 96.75 & 82.58 & -14.6 \\
    \addlinespace
    Emotion idle time, \%, \textdownarrow & 85.92 & 79.97 & -6.9 \\
    \addlinespace
    Silent animation, s & 0.36 & 0.65 & +80.8 \\
    \bottomrule
\end{tabularx}
\end{table}

The 75\% reduction in the number of interrupted actions primarily indicates that the baseline approach does not transfer well to a task with a long narrative, and secondly, it shows that the mechanisms in a multi-agent system work effectively. After all, interrupting actions leads to a "twitchy" behavior of the robot, which reduces naturalness.

The main advantage of the multi-agent system is an increase in the total number of actions by 50.9\% while reducing limb idleness. This indicates an increase in robot activity; according to research in \cite{Rodero2022}, the growth of gesture frequency contributes to better interaction with the interlocutor and the retention of attention. Despite the fact that this study provides values corresponding to the optimal number of gestures for a better perception of the speaker's speech, quantitative comparisons with those data were not made for the following reasons:
\begin{enumerate}{}{}
\item{The study focuses on hand gestures as micro-actions that complement speech (like tapping your finger to the rhythm of speech). In the case of a robot, where actions represent a sequence of kinematic actions of different limbs, a direct comparison is incorrect.}
\item{The correspondence of the number of gestures with the tempo of speech shown in the work is based on Spanish (169 words/min), which is very different in tempo from Russian (about 127 words/min) \cite{Bradlow2017}.}
\end{enumerate}

In addition to the increase in the total number of actions, it is worth noting the share of growth in non-default emotions (+42.9\%) and non-default body movements (+25.7\%). This indicates an increase in the expressiveness of non-verbal signals. At the same time, it is worth noting that returning to a neutral position is important for natural perception, especially after performing prominent, accent actions. That is why the multi-agent system shows a significant increase in the time of silent animation, because at the end of each scenario, the robot moves all the limbs to a neutral position.

\subsection{Experimental Conclusions}

The baseline approach focuses on generating behavior for short responses, where there are no conflicts between actions. However, in the long-term interaction task, this system demonstrates insufficient stability and richness of behavior. In contrast, the proposed multi-agent system is designed for long-term narratives that require the coordination of multiple actuators. Thanks to the mechanisms for prioritization and conflict resolution, the process of generating scenarios can be fully automated using LLM. At the same time, the system successfully handles short phrases, which ensures its versatility.

\subsection{Demonstration}

To provide a clear and illustrative example of the system's capabilities, a demonstration video\footnote{Demonstration video available on YouTube: \href{https://youtu.be/LXqljZSa8ZM}{\texttt{https://youtube.com/robofob/mentor-1}}} is proposed that presents the robot performing a scenario generated by the system, \textit{using the current article} as an extended description. 

\textbf{Generated scenario:}

\begin{lstlisting}
Ah, the Mentor-1, a marvel of modern robotics and, dare I say, a glimpse into the future where robots like me might just replace your friendly neighborhood tour guide. (*@\textbf{<facial:angry>}@*) But fear not, I promise to keep the rise of the machines to a minimum! (*@\textbf{<anim:right\_arm;show\_space;1>}@*) Now, let's dive into what makes me tick. I am an anthropomorphic robot, designed to guide you through this fascinating robotics laboratory. (*@\textbf{<facial:joy>}@*) My body is equipped with expressive features like active eyes and brows, and arms with five degrees of freedom, perfect for those dramatic gestures that keep you engaged. (*@\textbf{<pose:head\_and\_arms;proud>}@*) My primary function is to interact naturally with you, using both verbal and nonverbal cues. Imagine me as your robotic docent, but with a twist - I generate my own tour scenarios using large language models! (*@\textbf{<facial:question>}@*) This means I can adapt my storytelling style to suit different audiences, whether you're a seasoned engineer or just someone who wandered in out of curiosity. (*@\textbf{<anim:both\_arms;stretch\_both\_ways;1>}@*) My control system is a hybrid marvel, combining multi-agent resource management with automatic scenario generation. (*@\textbf{<facial:joy>}@*) This allows me to perform multiple actions simultaneously without tripping over my own circuits. My scenarios are crafted in two stages: first, I create a stylized narrative, then I integrate action tags for gestures and expressions. (*@\textbf{<facial:joy>}@*) This ensures my behavior is as natural as a robot can get, without the awkward pauses or mechanical movements you might expect. And yes, I can even make eye contact - something that might make you wonder if I'm plotting world domination or just trying to connect with you on a more human level. (*@\textbf{<facial:question>}@*) But rest assured, my main goal is to make your tour informative and enjoyable. (*@\textbf{<anim:head;nodding;2>}@*) So, as we wrap up this segment, remember: while I might be a robot, I'm here to make your experience as engaging as possible. (*@\textbf{<facial:joy>}@*) Who knows, maybe one day you'll prefer your tours with a side of AI!
\end{lstlisting}

The generation was performed with the following parameters: length of about 1200 characters, humorous style and aimed at an adult audience immersed in the research topic. The video illustrates not only the generated narrative but also the actions of the robot, appropriate to the context, synchronized with speech, thereby demonstrating the system's ability to create a coherent, contextually relevant, and stylistically adapted story accompanied by appropriate interactive behavior.

\section{Conclusion}\label{sec13}

The study carried out showed that the integration of a multi-agent control architecture with automatic LLM-based scenario generation improves the flexibility and naturalness of the behavior of an anthropomorphic robot in long-term narration tasks. Unlike existing approaches primarily designed for short dialogue responses, the developed system specifically addresses the challenges of prolonged storytelling by fully automating scenario preparation, generating rich parameterized nonverbal behavior while efficiently managing resource conflicts between multiple actuators. 

The proposed system automatically generates parameterized narratives that seamlessly adapt to specific styles, audiences, and desired durations. It also automatically supports these narratives with non-verbal actions and behaviors that perfectly match the spoken text.

Thus, the proposed solution allows the complete automation of generating context-specific narratives and corresponding actions at designated points within the tour, ensuring a unique experience with each execution. The hybrid control system provides flexibility, adaptability, and scalability. The abstract implementation using the ROS framework allows the system to be adapted to most anthropomorphic conversational robots.

Experiments on the MENTOR-1 robot guide have confirmed a decrease in action interruptions, an increase in behavioral richness, and a minimization of limb idleness compared to basic methods. The system has been successfully operating in the laboratory for almost a year, generating unique scenarios for each tour. The architecture is extensible for reactive interactions (responding to speech/gestures a high-level behavior tree), which is a future area of research.

The proposed system thus offers a versatile foundation for creating emotionally engaging robots capable of sustained and adaptive interaction in educational and service applications.

\section*{Funding}
This work was supported by the state assignment of the National Research Center "Kurchatov Institute".

\bibliographystyle{unsrtnat}
\bibliography{references}  
%%% Uncomment this line and comment out the ``thebibliography'' section below to use the external .bib file (using bibtex) .


\end{document}